%% file: main.tex
\documentclass[10pt,twocolumn,letterpaper]{article}

\usepackage[pagenumbers]{cvpr}

\input{preamble}

\definecolor{cvprblue}{rgb}{0.21,0.49,0.74}
\usepackage[pagebackref,breaklinks,colorlinks,allcolors=cvprblue]{hyperref}

\newcommand{\myparagraph}[1]{\vspace{0pt}\noindent{\bf{#1}}~~}
\def\confName{CVPR}
\def\confYear{2027}

\title{INCISE: Training-Free \tul{I}\tul{n}-\tul{C}ontext Segmentat\tul{i}on of \tul{S}urgical Vid\tul{e}os}

\author{
Yifan Xu$^{1}$\quad
Bofang Zheng$^{1}$\quad
Guangyu Meng$^{1}$\quad
Nishchal Sapkota$^{1}$\\
Xianshi Ma$^{1}$\quad
Xu Cao$^{2}$\quad
Tianjiao Yu$^{3}$\quad
Danny Z. Chen$^{1}$\\[0.6em]
{\normalsize $^{1}$University of Notre Dame\qquad
$^{2}$PediaMed AI\qquad
$^{3}$University of Illinois at Urbana-Champaign}
}

\begin{document}
\maketitle
\input{fig/teaser}
\input{sec/0_abstract}
\input{sec/1_intro}
\input{sec/2_related}
\input{sec/3_method}
\input{sec/4_experiments}
\input{sec/5_conclusion}
{
    \small
    \bibliographystyle{ieeenat_fullname}
    \bibliography{main}
}

\end{document}

%% file: preamble.tex
\usepackage{cuted}

\usepackage{currfile}

\usepackage{caption}

\newcommand{\ours}{INCISE\xspace}
\newcommand{\tul}[1]{\texorpdfstring{\underline{#1}}{#1}}
\input{figures_preamble}

\usepackage{multirow}
\usepackage{colortbl}
\usepackage{tabularx}
\newcommand{\defrow}{\rowcolor{gray!15}}

\newcommand{\grey}[1]{\textcolor{bpGray}{#1}}

%% file: figures_preamble.tex
\usepackage{tikz}
\usetikzlibrary{positioning,arrows.meta,fit,calc,backgrounds,decorations.pathreplacing,shapes.geometric}

\definecolor{bpOurs}{RGB}{0,114,178}
\definecolor{bpPrior}{RGB}{213,94,0}
\definecolor{bpData}{RGB}{0,158,115}
\definecolor{bpAccent}{RGB}{230,159,0}
\definecolor{bpGray}{RGB}{110,110,110}
\definecolor{bpLight}{RGB}{238,238,238}
\colorlet{bpOursFill}{bpOurs!18}
\colorlet{bpPriorFill}{bpPrior!15}
\colorlet{bpDataFill}{bpData!15}

\tikzset{
  bp/base/.style   ={draw=black!70, line width=0.5pt, rounded corners=2pt, align=center,
                     font=\sffamily\scriptsize, minimum height=0.6cm, inner sep=3pt},
  bp/module/.style ={bp/base, fill=white},
  bp/frozen/.style ={bp/base, fill=bpLight, draw=bpGray, text=black!80},
  bp/ours/.style   ={bp/base, fill=bpOursFill, draw=bpOurs, line width=0.7pt},
  bp/prior/.style  ={bp/base, fill=bpPriorFill, draw=bpPrior},
  bp/data/.style   ={bp/base, fill=bpDataFill, draw=bpData},
  bp/todo/.style   ={bp/base, dashed, draw=bpGray, text=bpGray},
  bp/flow/.style   ={-{Latex[length=1.6mm]}, line width=0.6pt, black!75},
  bp/note/.style   ={font=\sffamily\tiny, text=bpGray, align=center},
  bp/panel/.style  ={font=\sffamily\footnotesize\bfseries, anchor=west},
  bp/frame/.style  ={draw=black!70, fill=white, line width=0.4pt},
}

%% file: fig/teaser.tex
\begin{strip}
\centering
\includegraphics[width=\textwidth]{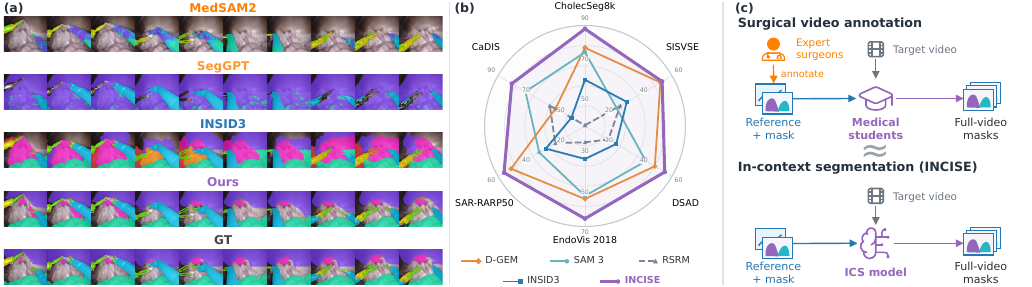}
\captionsetup{hypcap=false}
\captionof{figure}{\textbf{INCISE segments surgical videos in context without model training or target-video labels.} \textbf{(a)} Qualitative comparison across a surgical video sequence. \textbf{(b)} Test mIoU (\%) across six surgical benchmarks, where INCISE outperforms both fine-tuned models and methods given a target-video mask. \textbf{(c)} Just as surgeons annotate a few representative frames to guide medical students (top), INCISE selects references that cover the target video so a frozen model can segment it directly without training (bottom).}
\captionsetup{hypcap=true}
\label{fig:teaser}
\end{strip}

%% file: sec/0_abstract.tex
\begin{abstract}
Surgical video analysis requires video object segmentation (VOS), yet existing methods are commonly based on pretrained model fine-tuning using a large number of expert labels.
In-context segmentation (ICS) offers the promise of bypassing this high cost by segmenting a target video using a few annotated references, which are the sole source of surgical knowledge for a frozen model.
However, existing ICS methods tend to draw such references at random, which concentrate on a few views that dominate a surgical video and can miss other objects of interest (e.g., transient instruments and small structures).
To address these challenges, we propose \ours, a training-free ICS framework for surgical videos.
\ours selects unlabeled frames from multiple surgical videos that jointly cover the local appearances of a target video, and then segments the target video
using these references.
With only 1.7\% of training frames selected and annotated as references and no target video frame labeled, \ours obtains 67.2\% mean mIoU and outperforms the best SOTA fine-tuned method by 9.2 points.
We hope \ours can help establish ICS as a practical alternative to fine-tuning in surgical video analysis and other domains with a costly labeling burden.
\end{abstract}

%% file: sec/1_intro.tex
\section{Introduction}
\label{sec:intro}

Segmentation of instruments and anatomy in surgical videos supports both intraoperative guidance and surgical education.
Models trained on such masks delineate safe and dangerous dissection zones during an operation~\cite{madani2022artificial,protserov2024development}.
They also support automatic assessment of the critical view of safety in laparoscopic cholecystectomy~\cite{mascagni2025endoscapes}.
Such models further help teach surgery~\cite{kinoshita2024artificial,ward2021surgical}: medical students learned from a trained model to recognize anatomical landmarks nearly as well as from a surgeon tutor~\cite{ninomiya2026educational}.

\input{fig/paradigms}
Yet most existing methods fine-tune pretrained models on thousands of pixel-level masks, which
surgeons hardly have time to draw~\cite{maierhein2022surgical}.
To scale up annotation, clinicians typically annotate a few representative frames as visual guidelines and train non-expert annotators such as medical students to label the remaining frames~\cite{carstens2023dresden,grammatikopoulou2021cadis,skinner2024realtime}.
Even so, a medical student still needs about 25 minutes to annotate a single frame~\cite{scheikl2020deep}, so curating a training set remains a major bottleneck.
Fine-tuned video methods built on vision foundation models (VFMs) inherit this cost~\cite{ma2025medsam2,sapkota2026dgem}: they adapt a VFM on such labeled frames and further require an expert to label at least one frame of each target video (\cref{fig:paradigms}a).

In parallel, semi-supervised video object segmentation (Semi-VOS)~\cite{carion2026sam3,ravi2025sam2} removes the fine-tuning by keeping a VFM frozen, yet still needs expert masks on one or more frames of every target video (\cref{fig:paradigms}b).
Such masks are rarely available in practice, because an expert must annotate each new video before it can be segmented.
Even then, the labeled frames can miss classes that appear only elsewhere in the video, and choosing which frames to label is itself an open problem~\cite{griffin2019bubblenets,yin2021learning,delatolas2024learning}.

In-context segmentation (ICS) removes both costs: a frozen VFM segments a target video from a few annotated images of other videos (called references) given at test time, with neither fine-tuning nor a target-video mask~\cite{cuttano2026insid3,liu2025cosine,wang2023seggpt,zhang2024persam}.
Just as medical students learn what to segment from a few surgeon-labeled examples, an ICS model takes such references as context and segments the target surgical video itself (\cref{fig:teaser}c; \cref{fig:paradigms}c).
If a few annotated surgical frames could stand in for a training set, clinicians would need to annotate only these frames.
However, because these references are the sole source of surgical knowledge for the frozen VFM, existing ICS methods~\cite{sun2026rsrm,zhang2024persam} simply draw them at random~\cite{mousavi2025one}.

Choosing references at random may not work well since surgical appearances vary widely in duration and scale.
Specifically, major tissues and routine tools may remain visible in many frames, whereas a specialized instrument may enter the view only briefly~\cite{twinanda2016endonet} and a delicate structure may occupy only a small region~\cite{grammatikopoulou2021cadis,carstens2023dresden,psychogyios2023sarrarp50}.
Consequently, a handful of randomly drawn frames mostly capture frequent appearances~\cite{kang2019decoupling} and miss these brief or small classes altogether, thus leaving a frozen VFM with no reference to segment them.
Hence, we study whether a small budget of well-chosen references from other surgical videos can give a frozen VFM enough coverage to segment a target video.

To this end, we introduce \ours, a training-free approach that augments frozen VFMs with references selected for each target video (\cref{fig:teaser}a,b).
It operates in two stages.
(i) \emph{Coverage-based selection} picks a few unlabeled frames from other videos that jointly cover the cluster centers of the target video with DINOv3~\cite{simeoni2026dinov3} local features.
It weights the centers of rare clusters more, so brief instruments and small structures can also receive references.
(ii) \emph{Segmentation with independent pseudo-labels} reads the video frame by frame with DINOv3 and SAM~3 and labels each frame with these references.
Its memory keeps independent per-frame pseudo-labels instead of its own predictions, so errors do not accumulate. We make three contributions:
\begin{itemize}[nosep,leftmargin=*]
  \item We identify a core limitation of existing surgical video segmentation paradigms: supervised methods demand costly domain or per-video annotations, whereas training-free in-context approaches fail under complicated spatio-temporal distributions.
  \item We propose \ours, a training-free framework coupling coverage-based reference selection over weighted-rare local clusters with a history of independent pseudo-labels that prevents temporal error accumulation in frozen foundation models.
  \item Extensive evaluations across 6 surgical benchmarks and cross-domain transfers show that \ours surpasses fine-tuned paradigms and
  that strategic visual coverage outweighs raw annotation volume for robust generalization.
\end{itemize}

%% file: fig/paradigms.tex
\begin{figure}[t]
\centering
\includegraphics[width=\columnwidth]{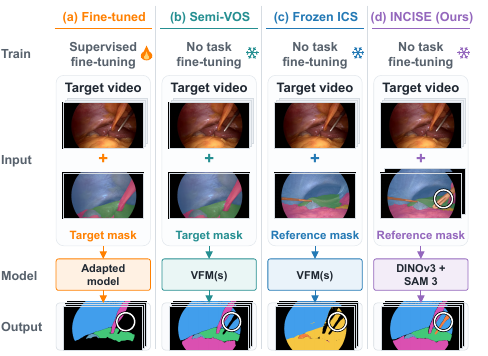}
\caption{\textbf{Training and input requirements.}
(a) Fine-tuned~\cite{ma2025medsam2,sapkota2026dgem}: supervised fine-tuning plus a target-video mask.
(b) Semi-VOS~\cite{carion2026sam3,ravi2025sam2}: a frozen VFM with such masks on one or more target frames.
(c) Frozen ICS~\cite{sun2026rsrm,zhang2024persam}: random external references, often missing brief classes.
(d) \ours: external references selected to cover the target video.
Circles mark a grasper,
a reference selected by \ours, which allows \ours to segment it in the target video.}
\label{fig:paradigms}
\end{figure}

%% file: sec/2_related.tex
\section{Related Work}
\label{sec:related}

\input{fig/overview}

\myparagraph{Surgical video segmentation.}
Surgical scene segmentation is benchmarked on laparoscopic, robotic and microscopic videos with pixel-level masks~\cite{hong2020cholecseg8k,allan2020robotic,psychogyios2023sarrarp50,grammatikopoulou2021cadis}.
Supervised models such as ISINet learn instrument classes from many annotated frames~\cite{gonzalez2020isinet}.
Recent methods adapt foundation models to medical and surgical video.
MedSAM and SurgicalSAM adapt SAM to medical and surgical images with mask supervision~\cite{ma2024medsam,yue2024surgicalsam}.
MedSAM2 fine-tunes SAM~2.1 on large collections of annotated 3D medical images and videos~\cite{ma2025medsam2}.
D-GEM trains a decoder on DINOv3 features and stores annotated frames of each target video in a dual memory~\cite{sapkota2026dgem}.
Both need masks drawn on frames of every target video, while \ours trains nothing and uses no mask of the target video.
Closest to our setting, Mousavi \etal~\cite{mousavi2025one} start SAM~2.1 in a cholecystectomy video from an annotated frame of another patient.
They draw this frame at random, report a large variance across draws and leave the choice of frame to future work.
\ours makes this choice for each target video.

\myparagraph{In-context segmentation.}
In-context segmentation models segment a target image from a few annotated references.
Generalist models learn this ability from segmentation data.
Painter and SegGPT train one model on mixtures of dense prediction and segmentation datasets~\cite{wang2023painter,wang2023seggpt}, and COSINE trains a decoder on frozen DINOv2 and CLIP features~\cite{liu2025cosine}.
UniverSeg and Tyche train such models on medical images~\cite{butoi2023universeg,rakic2024tyche}.
Training-free methods instead match the features of frozen VFMs.
PerSAM turns matches between the references and the target into point prompts for SAM~\cite{zhang2024persam}.
INSID3 segments with a single frozen DINOv3 backbone~\cite{cuttano2026insid3,simeoni2026dinov3}, and RSRM applies such matching to cross-domain few-shot segmentation~\cite{sun2026rsrm}.
All of these methods take the references as given.
On video, SegGPT, COSINE and PerSAM are prompted with annotations of the target video~\cite{wang2023seggpt,liu2025cosine,zhang2024persam}.
DC-SAM prompts SAM~2 from one external reference instead, but it trains a prompt generator and parts of SAM~2 for this task~\cite{qi2025dcsam}.

\myparagraph{Reference selection.}
Video object segmentation methods choose which frame of the target video to annotate~\cite{griffin2019bubblenets,yin2021learning,delatolas2024learning}, so every new video needs annotations that no other video can reuse.
Active learning selects samples to annotate for training a model~\cite{zheng2019representative,liu2023colossal,sener2018coreset}, and with few labels representative samples work better than uncertain ones~\cite{hacohen2022active}.
For a nearest-neighbor classifier, selecting training data reduces to maximizing a submodular facility-location function~\cite{wei2015submodularity}.
Visual in-context learning usually retrieves references for each target image from a fully annotated pool~\cite{zhang2023makes,wahd2025temporal}.
Vote-k, SCS and MVPS are closest to ours because they select from an unlabeled pool~\cite{su2023selective,suo2024rethinking,wu2024efficient}.
They work on text or single images, and SCS and MVPS train a selection agent or retriever on labeled data.
\ours trains nothing and picks unlabeled frames of other videos that jointly cover the weighted-rare cluster centers of each target video.

\myparagraph{Memory in video segmentation.}
Memory-based video object segmentation reads past frames together with their predicted masks~\cite{oh2019video,cheng2022xmem}.
SAM~2 propagates masks through a memory of recent frames encoded with its own predicted masks~\cite{ravi2025sam2}, and the SAM~3 tracker inherits this design~\cite{carion2026sam3}.
An erroneous mask in this memory is read again on later frames, so errors accumulate~\cite{ding2025sam2long,yang2024samurai}.
Later trackers filter what enters the memory by confidence, motion or distractor cues~\cite{carion2026sam3,ding2025sam2long,yang2024samurai,videnovic2025distractor}.
D-GEM likewise gates its predictions by confidence before writing them to memory~\cite{sapkota2026dgem}.
In all of them, outputs of the tracker itself still enter the memory.
Our history stores only pseudo-labels computed afresh from the references and the current frame, so no prediction is read back by later frames.

%% file: fig/overview.tex
\begin{figure*}[t]
\centering
\includegraphics[width=\textwidth]{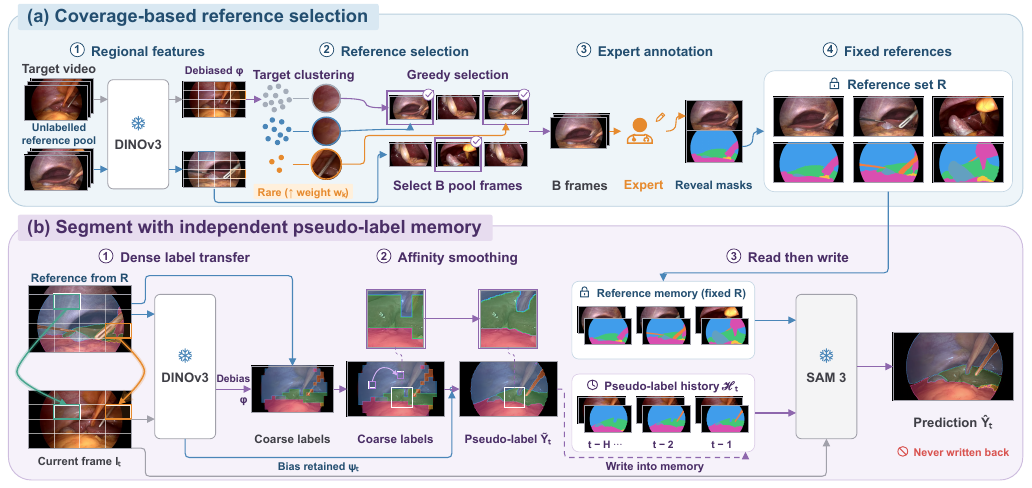}
\caption{\textbf{Overview of \ours.}
\textbf{(a) Coverage-based reference selection} (\cref{sec:selection}):
(1) Given a target video and an unlabeled pool, a frozen DINOv3 extracts debiased features $\phi$.
(2) We cluster the target features and greedily select $B$ pool frames that jointly cover the cluster centers, where the weights $w_k$ favor rare appearances (e.g., the orange cluster).
(3)--(4) Only these $B$ frames are labeled by experts to form a fixed reference set $R$.
\textbf{(b) Segmentation with independent pseudo-labels} (\cref{sec:segmenter}):
For each target frame $I_t$, (1) dense label transfer matches $\phi$ against $R$ to produce coarse labels, which (2) affinity smoothing refines into an independent pseudo-label $\tilde{Y}_t$ using the original features $\psi_t$.
(3) A frozen SAM~3 reads $R$ and the history $\mathcal{H}_t$ of the last $H$ pseudo-labels to predict $\hat{Y}_t$ \emph{before} $\tilde{Y}_t$ is written to $\mathcal{H}_t$.}
\label{fig:overview}
\end{figure*}

%% file: sec/3_method.tex
\section{Method}
\label{sec:method}

Our goal is to segment a target surgical video with $B$ annotated references alone, using frozen VFMs without training.
The references come from other videos of the same dataset and are chosen before any of their masks is revealed.
Surgical masks cost expert time, so only the chosen frames are ever annotated.
A key property of DINOv3 is the strong local self-similarity of its dense features~\cite{cuttano2026insid3}.
We use it to find suitable references: the local appearances of the target video group into clusters, and we select the pool frames that jointly cover their centers.
To keep errors from accumulating in memory, we derive pseudo-labels from dense correspondences between each frame and these references.
A frozen SAM~3 tracker stores these pseudo-labels in its memory instead of its own predictions.

Our approach, named \ours and illustrated in \cref{fig:overview}, first selects references from images alone so that they cover the local appearances of the target video (\cref{sec:selection}).
It then segments the video frame by frame from these references (\cref{sec:segmenter}).
Frozen DINOv3 features transfer labels, and a frozen SAM~3 tracker reads them with past pseudo-labels.

\myparagraph{Task definition.}
\label{sec:setting}
Each dataset provides a reference pool $\mathcal{P}=\{J_j\}_{j=1}^{N}$, the $N$ frames of its training split.
Their masks $M_j$ over $C$ foreground classes and the background are hidden at first.
This mirrors surgical practice, where recordings of similar procedures are plentiful but rarely annotated.
A target video $V=(I_1,\dots,I_T)$ comes from a different surgery than every frame in $\mathcal{P}$.
A selection rule summarizes the appearances of the unlabeled target video with image features alone and selects a reference set $R$ of $B$ pool frames.
Only then are the masks $\{M_r\}_{r\in R}$ revealed.
Conditioned on this fixed reference set, the segmenter reads $V$ once in temporal order and predicts each frame $\hat{Y}_t$ from $I_{1:t}$ and the annotated references.
The same reference set serves all classes.
Selecting $R$ over the full unlabeled video suits post-operative analysis, surgical education, and dataset curation; once $R$ is fixed (or reused from prior procedures), the segmenter itself is strictly causal and streams $V$ online in a single pass.

Once annotated, a pool frame serves every target video that selects it.
The budget $B$ counts references per target video, and methods are compared at equal $B$.
The dataset cost $\big|\bigcup_{V\in\mathcal{V}}R(V)\big|$ counts the distinct frames annotated for all evaluated target videos $\mathcal{V}$.

\subsection{Coverage-based reference selection}
\label{sec:selection}

\input{fig/cover}

Existing ICS methods draw references at random~\cite{mousavi2025one} or retrieve frames by global feature similarity~\cite{zhang2023makes,wahd2025temporal}.
Repeated anatomy can dominate retrieval, leaving brief instruments or small structures without reference labels.
In \cref{fig:cover}, none of the ten references retrieved by global similarity contains the suction tool of a SAR-RARP50 target video.
Coverage selects one at the lower left and recovers the tool (supplementary material).

We thus select references to cover the target video's local appearances, such as instrument parts or organ surfaces.

\myparagraph{Target clusters.}
Let $\phi(I)$ be the debiased DINOv3 patch features of a frame $I$ (\cref{sec:segmenter}).
Average pooling over a $G\times G$ grid and normalizing each cell gives $G^2$ regional features $z_g(I)$, reducing the cost of clustering all target frames.
Spherical $k$-means yields $K$ centers $c_k$ with assignment fractions $p_k$.
Each frame counts equally, so $p_k$ measures an appearance's prevalence.
We weight each center by
\begin{equation}
  w_k=\frac{p_k^{\beta}}{\sum_{j}p_j^{\beta}},
  \label{eq:weights}
\end{equation}
where $\beta\in[0,1]$ balances common and rare clusters.
$\beta=1$ keeps observed frequencies and favors dominant views; $\beta=0$ weights clusters equally but overweights outliers.

\myparagraph{Coverage objective.}
A pool frame covers a cluster center as well as its best-matching region does:
\begin{equation}
  s_k(J_j)=\tfrac{1}{2}\Big(1+\max_{g}\,\langle c_k,z_g(J_j)\rangle\Big)\in[0,1].
  \label{eq:cover}
\end{equation}
A reference set covers the target by the weighted best coverage of each cluster center:
\begin{equation}
  F(R)=\sum_{k=1}^{K} w_k \max_{j\in R} s_k(J_j).
  \label{eq:objective}
\end{equation}
We maximize $F$ greedily: starting from an empty set, we add the pool frame with the largest gain until $|R|=B$.
Repeated appearances add little; improving a center's coverage contributes its weight times the increase in similarity.

\myparagraph{Properties.}
$F$ is a monotone submodular facility-location function, so greedy selection attains at least a $(1-1/e)$ fraction of the optimal coverage~\cite{nemhauser1978analysis}.
Greedy selection gives nested reference sets, so annotators can follow the selected order and stop at any budget.
Selection uses no mask, and pool features are extracted once per dataset.

\subsection{Segmentation with independent pseudo-labels}
\label{sec:segmenter}

The segmenter combines two complementary sources.
Label transfer matches every patch of a frame against the references, so it finds each class wherever its appearance occurs but gives coarse boundaries.
SAM~3 memory attention gives sharp and temporally consistent masks, but it can only follow classes held in its memory.
We combine the two so that each covers the weakness of the other.

\myparagraph{Dense label transfer.}
DINOv3 features carry a positional bias, so patches at similar positions match even across unrelated images~\cite{cuttano2026insid3}.
Following INSID3, we remove the leading singular directions of a black image's features from the frozen DINOv3~\cite{simeoni2026dinov3} features $\psi$ and renormalize, which gives the debiased features $\phi$.
Label transfer works on individual DINOv3 patches rather than on the pooled grid cells used for selection.
Each reference patch $v$ carries the class distribution $y(v)$ of its mask pixels, since a $16\times16$ patch can span several classes.
A patch $u$ of frame $I_t$ takes the softmax-weighted labels of its $k$ most similar patches $\mathcal{N}_k(u)$ among all patches of the references:
\begin{equation}
  p_t(u)=\sum_{v\in\mathcal{N}_k(u)}
  \frac{e^{\langle\phi_t(u),\phi(v)\rangle/\tau}}
  {\sum_{v'\in\mathcal{N}_k(u)}e^{\langle\phi_t(u),\phi(v')\rangle/\tau}}\,y(v).
  \label{eq:knn}
\end{equation}
To make these estimates consistent within the frame, we smooth them over its patch affinity graph by label propagation~\cite{zhou2003learning}, a step we call affinity smoothing.
We build this graph from the original features $\psi$, which INSID3 also uses to group patches within an image.
Within one frame, their positional component helps group nearby patches of the same structure, while across images the same structure can appear anywhere.
Each patch links to its $k_d$ most similar patches $\mathcal{D}(u)$ of the same frame:
\begin{equation}
  q^{(s+1)}_t(u)=(1-\lambda)\,p_t(u)+\lambda\!\sum_{u'\in\mathcal{D}(u)}\!a_{uu'}\,q^{(s)}_t(u'),
  \label{eq:diffusion}
\end{equation}
where $a_{uu'}$ are softmax affinities and $q^{(0)}_t=p_t$.
After $L$ steps, upsampling $q^{(L)}_t$ and taking the most likely class gives the pseudo-label $\tilde{Y}_t$ of the frame.

\myparagraph{Reference memory and independent pseudo-labels.}
The SAM~3 tracker~\cite{carion2026sam3} encodes each reference with its mask as a conditioning frame, one object per class present in the references.
These memories are written once and never overwritten, so the references stay available for the whole video.
The pseudo-label $\tilde{Y}_t$ depends only on the references and the current frame.

\myparagraph{Pseudo-label history.}
Past frames provide context that no reference can: the same instruments, in the same light, a moment earlier.
For frame $t$, the tracker attends jointly to the reference memories and to the pseudo-labels of the last $H$ frames.
Reading the video once in temporal order with a bounded history matches the streaming trackers we compare against, and it keeps memory constant on long videos.
This read gives the prediction $\hat{Y}_t$.
The history stores $\tilde{Y}$, never $\hat{Y}$.
A prediction written back into memory would be read again by every later frame, so a single error could persist and spread (\cref{sec:ablation}).

%% file: fig/cover.tex
\begin{figure}[t]
\centering
\includegraphics[width=0.9\columnwidth]{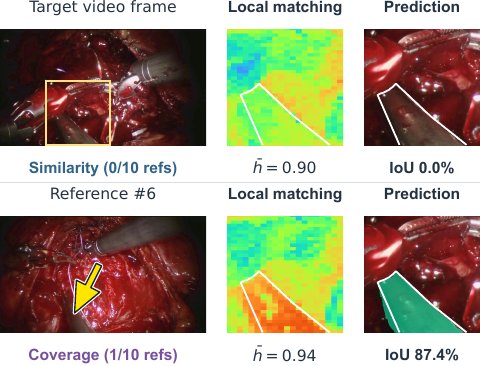}
\caption[Coverage-based selection recovers a tool missed by global similarity retrieval.]{\input{fig/insight_blind_spot_recovery/aligned_2x3_v2/caption}}
\label{fig:cover}
\end{figure}

%% file: fig/insight_blind_spot_recovery/aligned_2x3_v2/caption.tex
\textbf{Coverage-based selection recovers a tool missed by global similarity retrieval.}
\textbf{Left:} A target frame from SAR-RARP50 containing a suction tool (yellow box) and the external reference (\#6, yellow arrow) selected by our local coverage objective.
\textbf{Middle \& Right:} Local patch matching and segmentation predictions for global similarity retrieval (top, which selects 0/10 references containing the suction tool) versus our coverage selection (bottom, 1/10 references).
Heatmaps show the maximum reference patch similarity $h$; right panels show predicted masks (green), the ground truth (white contours).

%% file: sec/4_experiments.tex
\section{Experiments}
\label{sec:exp}
\input{tab/main}

We evaluate \ours on six surgical benchmarks against methods that receive a ground-truth mask of each target video.
We also compare with methods that, like ours, receive no mask of the target video and segment it from annotated references of other surgeries.
We report these results and the transfer to polyp segmentation in \cref{sec:main}.
We then analyze how references are chosen and how many are needed (\cref{sec:selection-exp}), and we ablate the segmenter (\cref{sec:ablation}).

\subsection{Experimental setup}
\label{sec:setup}

\myparagraph{Datasets.}
We evaluate on six multi-class surgical benchmarks from three specialties: general surgery (CholecSeg8k~\cite{hong2020cholecseg8k}, SISVSE~\cite{yoon2022surgical} and DSAD~\cite{carstens2023dresden}), urology (EndoVis 2018~\cite{allan2020robotic} and SAR-RARP50~\cite{psychogyios2023sarrarp50}) and ophthalmology (CaDIS~\cite{grammatikopoulou2021cadis}).
DSAD labels only organs and SAR-RARP50 only tools and suture material; the others label both anatomy and instruments.
Each training split forms the reference pool, and each test video is a target video (supplementary material).
PolypGen~\cite{ali2023polypgen} tests transfer to colonoscopy.

\myparagraph{Metrics.}
We report mIoU including background, except on CaDIS, which has no background class.
Class IoUs are averaged within each frame, omitting classes absent from both prediction and ground truth.
Frames are then averaged per target video, target videos per dataset, and the six datasets equally for the mean.
On PolypGen we report the frame-averaged Dice and IoU defined in their paper~\cite{ali2023polypgen}.

\myparagraph{Implementation and protocol.}
\ours uses DINOv3 ViT-L/16~\cite{simeoni2026dinov3} and the SAM~3 tracker~\cite{carion2026sam3}, both frozen.
Selection uses $K=128$ centers, $\beta=0.5$ and $B=10$ references per target video, and the segmenter keeps $H=12$ history frames.
One configuration serves every dataset.

\subsection{Main results}
\label{sec:main}

\myparagraph{Baselines.}
\Cref{tab:main} compares \ours with state-of-the-art surgical, medical and in-context segmentation methods in mIoU on the six surgical benchmarks.
We group the baselines by whether they train and whether they are given a mask of the target video.
The primary baselines are training-free ICS methods such as PerSAM~\cite{zhang2024persam}, RSRM~\cite{sun2026rsrm} and INSID3~\cite{cuttano2026insid3}, which share our setting.
They receive no mask of the target video and segment it from references of other surgeries drawn at random.
For RSRM and INSID3, we also report variants that receive the reference sets selected by \ours.
We also report methods that receive a ground-truth mask of each target video, such as the training-free trackers SAM~2~\cite{ravi2025sam2} and SAM~3~\cite{carion2026sam3}.
Fine-tuned methods such as MedSAM2~\cite{ma2025medsam2}, SegGPT~\cite{wang2023seggpt}, COSINE~\cite{liu2025cosine} and D-GEM~\cite{sapkota2026dgem} take the same mask as input and also train with mask supervision.
These methods need an expert to annotate every target video, so they operate under a stronger supervision regime than ours.
\ours trains nothing and uses no mask of the target video, and it alone selects its references to cover that video.

\input{fig/qualitative}

\myparagraph{Surgical video segmentation.}
\ours achieves the best result on all six benchmarks without training or any mask of the target video (\cref{tab:main}).
Its mean of 67.2\% is 9.2 points above the strongest fine-tuned method and 14.7 points above SAM~3 prompted with a mask of the target video.
\ours leads in each specialty, by 5.4 points in general surgery, 6.9 in urology and 6.1 in ophthalmology.
\Cref{fig:qualitative} shows qualitative results.

Methods given a mask of the target video, such as SegGPT and COSINE, can follow only the classes in that mask.
Classes that appear later are common in surgery: across the six benchmarks, 117 of the 370 classes of the target videos are absent from the given mask.
The references of \ours stay in memory for the whole video, so a class that enters late can still be matched.

Training-free ICS methods share our setting, yet \ours outperforms them by 26.9 points or more.
Their references come from random draws or similarity retrieval (\cref{sec:selection-exp}), and they segment each frame independently, without temporal information (\cref{sec:ablation}).
However, our selection helps RSRM and INSID3 (7.2 and 2.6 points in mean mIoU), as the selected references cover the appearances of the target video.
In contrast, \ours also uses temporal information through its pseudo-label history and stays 24.3 points ahead of both with the same references.
Across the six reference pools, \ours annotates 1.7\% of the pool frames (supplementary material).
Of its reference slots, 18.9\% reuse a frame already annotated for another target video.

\input{tab/polyp}

\myparagraph{Polyp segmentation.}
\label{sec:polyp}
PolypGen collects its data at six medical centres, and its benchmark tests whether a segmenter generalizes to a centre it has not seen.
Its supervised baselines are trained on five centres and tested on video sequences of the sixth.
\ours draws its references from the same five centres and keeps the configuration used for surgical video.
\ours reaches a Dice of 72.9\% (\cref{tab:polyp}), above the DeepLabV3+ and ResNet-UNet baselines.

\input{tab/selection}

\subsection{On the choice of references}
\label{sec:selection-exp}

Since the references are the only source of surgical knowledge for the frozen models, we ask which frames should be annotated and how many are needed.

\myparagraph{Coverage versus similarity.}
We first compare coverage with random references and with similarity retrieval in \cref{tab:selection}.
Similarity retrieval is the usual choice in visual in-context learning~\cite{zhang2023makes,wahd2025temporal}: it ranks pool frames by the cosine similarity of their global DINOv3 feature to the mean feature of the target video, which favors its dominant view.
Coverage improves every dataset with both segmenters, indicating that the gain originates in the references rather than in SAM~3.
With the full \ours it adds 5.9 points over similarity and 10.5 over random references, and with dense label transfer alone it adds 5.6 and 9.6.
As shown in \cref{fig:budget}b, performance stays stable across the balancing strength $\beta$, since clustering local features already enforces broad visual diversity; setting $\beta=0.5$ yields the best overall mean (70.3\%) and improves worst-frame coverage and tool-heavy benchmarks such as SAR-RARP50 (+0.5 points over $\beta=1$, supplementary material).

\input{fig/budget}

\myparagraph{Annotation budget and cost.}
\label{sec:budget}
Finally, we study the number of references per target video and its cost in \cref{fig:budget}c and d (per dataset in the supplementary material).
Accuracy rises steeply up to about ten references and then saturates, indicating that beyond a small budget the choice of frames matters more than their number.
Going from one to ten references raises the mean by 8.6 points, while more references add at most 0.17.
The coverage objective explains this saturation: it is submodular, so each added reference contributes less new coverage of the target video than the one before.
Notably, even at $B=1$—matching the single-mask budget of target-guided baselines—\ours reaches 60.7\% mean mIoU (supplementary material), already surpassing SAM~3~\cite{carion2026sam3} (55.9\%) and SegGPT~\cite{wang2023seggpt} (48.6\%) prompted with the ground-truth first-frame mask of the target video.
We also give the target-guided baselines the masks of the first ten frames of each target video, close to the 8.1 pool frames that \ours annotates per target video on average.
\ours still exceeds the mean mIoU of SAM~3 by 6.6 points and of SAM~2 by 7.7 points (supplementary material).
Its margins over D-GEM and MedSAM2 are 12.6 and 25.9 points.
From 10 to 100 references, peak GPU memory rises from 7.46 to 33.90\,GB and per-frame time from 0.95 to 2.54\,s.
We therefore use ten references per target video.
\input{tab/ablation}

\subsection{Ablation study}
\label{sec:ablation}

To understand how the segmenter turns the selected references into masks, we ablate each of its components in \cref{tab:ablation} and its history length in \cref{fig:budget}a.

\myparagraph{Dense label transfer.}
We first study the dense label transfer stage that produces each frame's pseudo-label.
Without feature debiasing, patches at similar spatial positions spuriously match even across unrelated frames; without affinity smoothing, each patch is classified in isolation.
Adding debiasing raises the mean from 58.7 to 61.2, and affinity smoothing raises it further to 64.1, confirming that positional bias and isolated patch noise both limit label transfer.
The remaining label-transfer parameters change the mean by at most 0.2 points, so we keep them fixed.

\myparagraph{Pseudo-label history and reference memory.}
We next add the two memory banks that SAM~3 attends to: the pseudo-label history provides temporal context from the current video, while the reference memory provides annotated masks of every selected class.
Conditioning on the pseudo-labels of past frames adds 3.0 points to label transfer, and the reference memory adds another 3.2.

\myparagraph{Independent pseudo-labels.}
We then examine what the temporal memory should store.
Video trackers (e.g., SAM~3) recursively write their own decoded predictions back into memory, allowing a single frame's error to persist across all subsequent frames; our pseudo-labels instead depend only on the references and the current frame.
Replacing our independent pseudo-labels with recursive predictions hurts every dataset and drops the mean by 3.3 points, no better than removing the reference memory.

\myparagraph{History length.}
Finally, we vary the number of past frames $H$ retained for SAM~3 in \cref{fig:budget}a.
Removing the pseudo-label history ($H=0$, which reduces to prompting SAM~3 with the selected reference memories alone) drops the mean by 6.1 points, and a single past frame already recovers 4.1 of them.
Accuracy rises steeply and saturates around $H=12$, changing by less than 0.2 points between 12 and 48 frames, which indicates that most temporal context comes from the most recent frames.
We therefore keep $H=12$ history frames.

\subsection{Cross-dataset transfer}
\label{sec:transfer}

Finally, we test whether \ours can segment a target video with references from another dataset (\cref{tab:transfer}).
The source pools transfer within a procedure (Endoscapes~\cite{mascagni2025endoscapes} and EndoVis 2017~\cite{allan2019endovis2017}), across procedures (DSAD~\cite{carstens2023dresden}) and from synthetic to real images (Synth-Colon~\cite{moreu2022synthetic}).
Each pool holds all eligible source images, and we evaluate on the classes shared with the target.

\input{tab/cross_dataset}

Coverage gives the best mIoU in all four pairs, with the largest gain of 7.66 points over the stronger baseline on Endoscapes $\to$ CholecSeg8k.
Under the appearance shifts of EndoVis 2017 $\to$ 2018 and Synth-Colon $\to$ PolypGen C6, similarity retrieval even falls below random selection.
Matching the mean target feature then selects redundant source frames of a single visual mode, which coverage avoids by covering the whole target video.
These results show that \ours can reuse existing annotated datasets, including synthetic renders, for new surgical procedures and clinical centers without target-domain annotations.

%% file: tab/main.tex
\begin{table*}[t]
\centering
\caption{\textbf{Comparison with state-of-the-art methods on six surgical benchmarks (mIoU in \%~, $\uparrow$).}
Our evaluations exclude the first frame of each target video.
\emph{Encoder}: image backbone.
\emph{Fine-tuned}: trained with mask supervision for in-context, medical or surgical segmentation; \emph{training-free}: pretrained models used as released.
\emph{+ our refs.}: the baseline receives the reference sets selected by \ours.
\grey{Gray}: trained on the training split of the corresponding dataset.
Best result in \textbf{bold}; second best \underline{underlined}.}
\label{tab:main}
\footnotesize
\setlength{\tabcolsep}{3.5pt}
\resizebox{\textwidth}{!}{%
\begin{tabular}{ll|ccc|cc|c|c}
\toprule
& & \multicolumn{3}{c|}{\textbf{General surgery}} & \multicolumn{2}{c|}{\textbf{Urology}} & \multicolumn{1}{c|}{\textbf{Ophthalmology}} & \\
\cmidrule(lr){3-5}\cmidrule(lr){6-7}\cmidrule(lr){8-8}
\textbf{Method} & \textbf{Encoder} & \textbf{CholecSeg8k} & \textbf{SISVSE} & \textbf{DSAD} & \textbf{EndoVis 2018} & \textbf{SAR-RARP50} & \textbf{CaDIS} & \textbf{Mean} \\
\midrule
\multicolumn{9}{@{}l}{\textbf{Fine-tuned}: \emph{mask of the target video given}} \\
\hspace{0.5em}MedSAM2~\cite{ma2025medsam2} & Hiera & \underline{86.9} & 16.1 & 23.5 & 17.0 & 23.5 & 43.4 & 35.1 \\
\hspace{0.5em}SegGPT~\cite{wang2023seggpt} & ViT & 85.7 & 16.5 & 30.6 & 39.7 & 20.3 & 60.1 & 42.1 \\
\hspace{0.5em}COSINE~\cite{liu2025cosine} & DINOv2 + CLIP & 76.3 & 15.9 & 29.7 & 32.5 & 22.1 & 39.5 & 36.0 \\
\hspace{0.5em}D-GEM~\cite{sapkota2026dgem} & DINOv3 & \grey{78.9} & \grey{\underline{53.3}} & \grey{\underline{50.1}} & \grey{\underline{56.3}} & \grey{\underline{52.5}} & \grey{57.0} & \grey{\underline{58.0}} \\
\midrule
\multicolumn{9}{@{}l}{\textbf{Training-free}: \emph{mask of the target video given}} \\
\hspace{0.5em}SAM~2~\cite{ravi2025sam2} & Hiera & 72.4 & 24.2 & 44.9 & 50.1 & 33.6 & \underline{76.0} & 50.2 \\
\hspace{0.5em}SAM~3~\cite{carion2026sam3} & PE & 76.7 & 28.1 & 44.9 & 54.8 & 36.3 & 74.2 & 52.5 \\
\midrule
\multicolumn{9}{@{}l}{\textbf{Training-free}: \emph{no mask of the target video}} \\
\hspace{0.5em}PerSAM~\cite{zhang2024persam} & SAM & 35.2 & 12.4 & 20.9 & 23.4 & 12.6 & 30.8 & 22.5 \\
\hspace{0.5em}RSRM~\cite{sun2026rsrm} & DINOv3 & 40.4 & 30.3 & 23.4 & 28.1 & 27.1 & 60.2 & 34.9 \\
\hspace{0.5em}\rotatebox[origin=c]{180}{$\Lsh$} \emph{+ our refs.} & DINOv3 & 59.2 & 31.6 & 30.9 & 38.2 & 31.5 & 61.1 & 42.1 \\
\hspace{0.5em}INSID3~\cite{cuttano2026insid3} & DINOv3 & 63.0 & 34.0 & 27.9 & 36.4 & 32.5 & 47.9 & 40.3 \\
\hspace{0.5em}\rotatebox[origin=c]{180}{$\Lsh$} \emph{+ our refs.} & DINOv3 & 70.8 & 35.1 & 27.8 & 41.2 & 32.3 & 50.2 & 42.9 \\
\defrow \hspace{0.5em}\textbf{\ours} \emph{(ours)} & DINOv3 + SAM~3 & \textbf{88.3} & \textbf{54.3} & \textbf{55.8} & \textbf{66.1} & \textbf{56.5} & \textbf{82.1} & \textbf{67.2} \\
\bottomrule
\end{tabular}
}
\end{table*}

%% file: fig/qualitative.tex
\begin{figure*}[t]
\centering
\includegraphics[width=\textwidth]{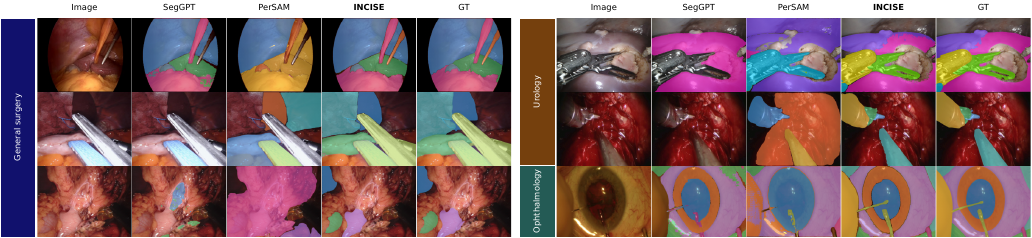}
\caption{\textbf{Qualitative comparison on six surgical benchmarks.}
Datasets are grouped as in \cref{tab:main}.}
\label{fig:qualitative}
\end{figure*}

%% file: tab/polyp.tex
\begin{table}[t]
\centering
\caption{\textbf{Transfer to polyp segmentation on PolypGen.}
\ours uses their frames only as references and is never trained on them. Baseline values are reported by~\cite{ali2023polypgen} and rounded there. \ours keeps the surgical configuration fixed.}
\label{tab:polyp}
\footnotesize
\begin{tabularx}{\linewidth}{Xccc}
\toprule
Method & Training & Dice & IoU \\
\midrule
ResNet-UNet~\cite{ronneberger2015unet,he2016resnet} & supervised & 70 & 65 \\
DeepLabV3+~\cite{chen2018encoder} & supervised & 71 & 65 \\
\defrow \textbf{\ours} \emph{(ours)}& none & \textbf{72.9} & \textbf{67.5} \\
\bottomrule
\end{tabularx}
\end{table}

%% file: tab/selection.tex
\begin{table}[t]
\centering
\caption{\textbf{Effect of reference selection.} mIoU (\%) of coverage, random and similarity references on CholecSeg8k (Chol.), EndoVis 2018 (EV18) and SAR-RARP50 (SAR).
\emph{Random}: frames drawn uniformly from the pool.
\emph{Similarity}: the frames whose global DINOv3 feature is most similar to the mean feature of the target video.
All rows use the earlier configuration (supplementary material).}
\label{tab:selection}
\footnotesize
\setlength{\tabcolsep}{2.5pt}
\begin{tabularx}{\linewidth}{Xcccc}
\toprule
\textbf{Selection} & \textbf{Chol.} & \textbf{EV18} & \textbf{SAR} & \textbf{Mean} \\
\midrule
\multicolumn{5}{l}{\emph{Dense label transfer}} \\
\hspace{0.5em}random & 68.5 & 48.0 & 43.3 & 53.3 \\
\hspace{0.5em}similarity & 70.4 & 56.2 & 45.2 & 57.3 \\
\defrow \hspace{0.5em}coverage (ours) & \textbf{80.9} & \textbf{60.6} & \textbf{47.1} & \textbf{62.8} \\
\cmidrule[\lightrulewidth](l{\dimexpr\tabcolsep+0.5em\relax}){1-5}
\multicolumn{5}{l}{\emph{Full \ours}} \\
\hspace{0.5em}random & 78.0 & 48.8 & 49.8 & 58.9 \\
\hspace{0.5em}similarity & 79.6 & 58.1 & 52.5 & 63.4 \\
\defrow \hspace{0.5em}coverage (ours) & \textbf{87.1} & \textbf{65.3} & \textbf{55.6} & \textbf{69.3} \\
\bottomrule
\end{tabularx}
\end{table}

%% file: fig/budget.tex
\begin{figure*}[t]
\centering
\includegraphics[width=0.8\textwidth]{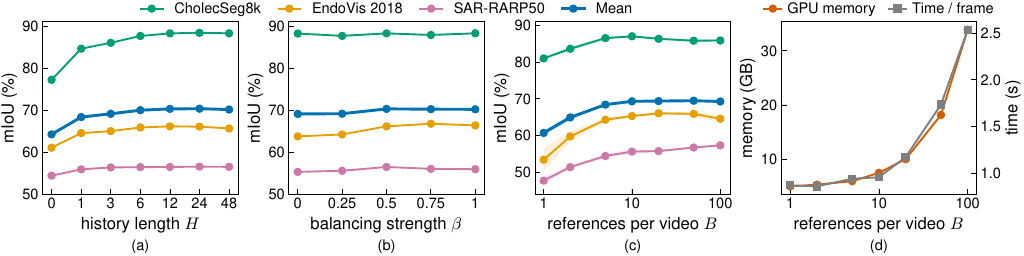}
\caption{\textbf{History length, cluster balancing, annotation budget and computational cost.}
mIoU against (a)~the history length $H$ and (b)~balancing strength $\beta$.
(c)~mIoU against reference number per target video.
(d)~Peak GPU memory in GB and time per frame (seconds).}
\label{fig:budget}
\end{figure*}

%% file: tab/ablation.tex
\begin{table}[t]
\centering
\caption{\textbf{Effect of segmenter components.} We ablate the segmenter on CholecSeg8k (Chol.), EndoVis 2018 (EV18) and SAR-RARP50 (SAR) in mIoU (\%).}
\label{tab:ablation}
\footnotesize
\setlength{\tabcolsep}{2.5pt}
\begin{tabularx}{\linewidth}{Xcccc}
\toprule
\textbf{Variant} & \textbf{Chol.} & \textbf{EV18} & \textbf{SAR} & \textbf{Mean} \\
\midrule
\multicolumn{5}{l}{\emph{Adding components}} \\
\hspace{0.5em}label transfer & 74.6 & 55.9 & 45.4 & 58.7 \\
\hspace{0.5em}+ debiasing & 77.2 & 58.8 & 47.7 & 61.2 \\
\hspace{0.5em}+ affinity smoothing & 83.1 & 62.2 & 46.9 & 64.1 \\
\hspace{0.5em}+ pseudo-label history & 84.4 & \textbf{66.9} & 49.9 & 67.1 \\
\defrow \hspace{0.5em}+ reference memory (full model) & \textbf{88.3} & 66.1 & \textbf{56.5} & \textbf{70.3} \\
\cmidrule[\lightrulewidth](l{\dimexpr\tabcolsep+0.5em\relax}){1-5}
\multicolumn{5}{l}{\emph{Full model}} \\
\hspace{0.5em}recursive history & 83.5 & 62.4 & 55.1 & 67.0 \\
\bottomrule
\end{tabularx}
\end{table}

%% file: tab/cross_dataset.tex
\begin{table}[t]
\centering
\caption{\textbf{Cross-dataset transfer (mIoU \%).}
Best in \textbf{bold}.}
\label{tab:transfer}
\footnotesize
\setlength{\tabcolsep}{2pt}
\renewcommand{\arraystretch}{1.1}
\begin{tabular*}{\linewidth}{@{\extracolsep{\fill}}lrrr@{}}
\toprule
\textbf{Reference $\rightarrow$ Target} & \textbf{Random} & \textbf{Similarity} & \textbf{Coverage} \\
\midrule
\multicolumn{4}{@{}l}{\emph{Within-procedure}} \\
\hspace{0.5em}Endoscapes $\rightarrow$ CholecSeg8k & 39.38 & 55.43 & \textbf{63.09} \\
\hspace{0.5em}EndoVis 2017 $\rightarrow$ 2018 & 63.61 & 60.62 & \textbf{69.12} \\
\midrule
\multicolumn{4}{@{}l}{\emph{Cross-procedure}} \\
\hspace{0.5em}DSAD $\rightarrow$ CholecSeg8k & 51.93 & 81.43 & \textbf{83.34} \\
\midrule
\multicolumn{4}{@{}l}{\emph{Synthetic-to-real}} \\
\hspace{0.5em}Synth-Colon $\rightarrow$ PolypGen C6 & 76.83 & 76.52 & \textbf{78.51} \\
\bottomrule
\end{tabular*}
\end{table}

%% file: sec/5_conclusion.tex
\section{Conclusion}
\label{sec:conclusion}

We showed that in-context segmentation can be an alternative to fine-tuning for surgical video once its references are chosen with care.
\ours selects a few frames of other surgeries that jointly cover the local appearances of each target video.
A frozen tracker then reads these references with a history of independent pseudo-labels, so errors do not accumulate.
Without training or any mask of the target video, it outperforms methods fine-tuned on thousands of surgical frames.
Beyond a small budget, which frames are annotated matters more than how many.
We hope this work makes ICS practical in surgery and other domains with costly labels.

%% file: main.bbl
\begin{thebibliography}{61}
\providecommand{\natexlab}[1]{#1}
\providecommand{\url}[1]{\texttt{#1}}
\expandafter\ifx\csname urlstyle\endcsname\relax
  \providecommand{\doi}[1]{doi: #1}\else
  \providecommand{\doi}{doi: \begingroup \urlstyle{rm}\Url}\fi

\bibitem[Ali et~al.(2023)Ali, Jha, Ghatwary, Realdon, Cannizzaro, Salem, Lamarque, Daul, Riegler, Anonsen, et~al.]{ali2023polypgen}
Sharib Ali, Debesh Jha, Noha Ghatwary, Stefano Realdon, Renato Cannizzaro, Osama~E Salem, Dominique Lamarque, Christian Daul, Michael~A Riegler, Kim~V Anonsen, et~al.
\newblock A multi-centre polyp detection and segmentation dataset for generalisability assessment.
\newblock \emph{Scientific Data}, 10\penalty0 (1):\penalty0 75, 2023.

\bibitem[Allan et~al.(2019)Allan, Shvets, Kurmann, Zhang, Duggal, Su, Rieke, Laina, Kalavakonda, Bodenstedt, et~al.]{allan2019endovis2017}
Max Allan, Alex Shvets, Thomas Kurmann, Zichen Zhang, Rahul Duggal, Yun-Hsuan Su, Nicola Rieke, Iro Laina, Niveditha Kalavakonda, Sebastian Bodenstedt, et~al.
\newblock 2017 robotic instrument segmentation challenge.
\newblock \emph{arXiv preprint arXiv:1902.06426}, 2019.

\bibitem[Allan et~al.(2020)Allan, Kondo, Bodenstedt, Leger, Kadkhodamohammadi, Luengo, Fuentes, Flouty, Mohammed, Pedersen, et~al.]{allan2020robotic}
Max Allan, Satoshi Kondo, Sebastian Bodenstedt, Stefan Leger, Rahim Kadkhodamohammadi, Imanol Luengo, Felix Fuentes, Evangello Flouty, Ahmed Mohammed, Marius Pedersen, et~al.
\newblock 2018 robotic scene segmentation challenge.
\newblock \emph{arXiv preprint arXiv:2001.11190}, 2020.

\bibitem[Butoi et~al.(2023)Butoi, Ortiz, Ma, Sabuncu, Guttag, and Dalca]{butoi2023universeg}
Victor~Ion Butoi, Jose Javier~Gonzalez Ortiz, Tianyu Ma, Mert~R Sabuncu, John Guttag, and Adrian~V Dalca.
\newblock {UniverSeg}: Universal medical image segmentation.
\newblock In \emph{2023 IEEE/CVF international conference on computer vision (ICCV)}, pages 21381--21394. IEEE, 2023.

\bibitem[Carion et~al.(2026)Carion, Gustafson, Hu, Debnath, Hu, Suris Coll-Vinent, Ryali, Alwala, Khedr, Huang, et~al.]{carion2026sam3}
Nicolas Carion, Laura Gustafson, Yuan-Ting Hu, Shoubhik Debnath, Ronghang Hu, Didac Suris Coll-Vinent, Chaitanya Ryali, Kalyan~Vasudev Alwala, Haitham Khedr, Andrew Huang, et~al.
\newblock {SAM}~3: Segment anything with concepts.
\newblock In \emph{International conference on learning representations}, pages 138846--138923, 2026.

\bibitem[Carstens et~al.(2023)Carstens, Rinner, Bodenstedt, Jenke, Weitz, Distler, Speidel, and Kolbinger]{carstens2023dresden}
Matthias Carstens, Franziska~M Rinner, Sebastian Bodenstedt, Alexander~C Jenke, J{\"u}rgen Weitz, Marius Distler, Stefanie Speidel, and Fiona~R Kolbinger.
\newblock The {Dresden} surgical anatomy dataset for abdominal organ segmentation in surgical data science.
\newblock \emph{Scientific Data}, 10\penalty0 (1):\penalty0 3, 2023.

\bibitem[Chen et~al.(2018)Chen, Zhu, Papandreou, Schroff, and Adam]{chen2018encoder}
Liang-Chieh Chen, Yukun Zhu, George Papandreou, Florian Schroff, and Hartwig Adam.
\newblock Encoder-decoder with atrous separable convolution for semantic image segmentation.
\newblock In \emph{European conference on computer vision}, pages 833--851. Springer, 2018.

\bibitem[Cheng and Schwing(2022)]{cheng2022xmem}
Ho~Kei Cheng and Alexander~G Schwing.
\newblock {XMem}: Long-term video object segmentation with an {Atkinson-Shiffrin} memory model.
\newblock In \emph{European conference on computer vision}, pages 640--658. Springer, 2022.

\bibitem[Cuttano et~al.(2026)Cuttano, Trivigno, Reich, Cremers, Masone, and Roth]{cuttano2026insid3}
Claudia Cuttano, Gabriele Trivigno, Christoph Reich, Daniel Cremers, Carlo Masone, and Stefan Roth.
\newblock {INSID3}: Training-free in-context segmentation with {DINOv3}.
\newblock In \emph{Third Workshop on Visual Concepts}, 2026.

\bibitem[Delatolas et~al.(2024)Delatolas, Kalogeiton, and Papadopoulos]{delatolas2024learning}
Thanos Delatolas, Vicky Kalogeiton, and Dim~P Papadopoulos.
\newblock Learning the what and how of annotation in video object segmentation.
\newblock In \emph{2024 IEEE/CVF Winter Conference on Applications of Computer Vision (WACV)}, pages 6936--6946. IEEE, 2024.

\bibitem[Ding et~al.(2025)Ding, Qian, Dong, Zhang, Zang, Cao, Guo, Lin, and Wang]{ding2025sam2long}
Shuangrui Ding, Rui Qian, Xiaoyi Dong, Pan Zhang, Yuhang Zang, Yuhang Cao, Yuwei Guo, Dahua Lin, and Jiaqi Wang.
\newblock {SAM2Long}: Enhancing {SAM}~2 for long video segmentation with a training-free memory tree.
\newblock In \emph{2025 IEEE/CVF International Conference on Computer Vision (ICCV)}, pages 13614--13624. IEEE, 2025.

\bibitem[Gonz{\'a}lez et~al.(2020)Gonz{\'a}lez, Bravo-S{\'a}nchez, and Arbelaez]{gonzalez2020isinet}
Cristina Gonz{\'a}lez, Laura Bravo-S{\'a}nchez, and Pablo Arbelaez.
\newblock {ISINet}: an instance-based approach for surgical instrument segmentation.
\newblock In \emph{International conference on medical image computing and computer-assisted intervention}, pages 595--605. Springer, 2020.

\bibitem[Grammatikopoulou et~al.(2021)Grammatikopoulou, Flouty, Kadkhodamohammadi, Quellec, Chow, Nehme, Luengo, and Stoyanov]{grammatikopoulou2021cadis}
Maria Grammatikopoulou, Evangello Flouty, Abdolrahim Kadkhodamohammadi, Gwenol{\'e} Quellec, Andre Chow, Jean Nehme, Imanol Luengo, and Danail Stoyanov.
\newblock {CaDIS}: {Cataract} dataset for surgical {RGB}-image segmentation.
\newblock \emph{Medical Image Analysis}, 71:\penalty0 102053, 2021.

\bibitem[Griffin and Corso(2019)]{griffin2019bubblenets}
Brent~A Griffin and Jason~J Corso.
\newblock {BubbleNets}: Learning to select the guidance frame in video object segmentation by deep sorting frames.
\newblock In \emph{2019 IEEE/CVF Conference on Computer Vision and Pattern Recognition (CVPR)}, pages 8906--8915. IEEE, 2019.

\bibitem[Hacohen et~al.(2022)Hacohen, Dekel, and Weinshall]{hacohen2022active}
Guy Hacohen, Avihu Dekel, and Daphna Weinshall.
\newblock Active learning on a budget: Opposite strategies suit high and low budgets.
\newblock \emph{arXiv preprint arXiv:2202.02794}, 2022.

\bibitem[He et~al.(2016)He, Zhang, Ren, and Sun]{he2016resnet}
Kaiming He, Xiangyu Zhang, Shaoqing Ren, and Jian Sun.
\newblock Deep residual learning for image recognition.
\newblock In \emph{Proceedings of the IEEE conference on computer vision and pattern recognition}, pages 770--778, 2016.

\bibitem[Hong et~al.(2020)Hong, Kao, Kuo, Wang, Chang, and Shih]{hong2020cholecseg8k}
W-Y Hong, C-L Kao, Y-H Kuo, J-R Wang, W-L Chang, and C-S Shih.
\newblock {CholecSeg8k}: a semantic segmentation dataset for laparoscopic cholecystectomy based on {Cholec80}.
\newblock \emph{arXiv preprint arXiv:2012.12453}, 2020.

\bibitem[Kang et~al.(2019)Kang, Xie, Rohrbach, Yan, Gordo, Feng, and Kalantidis]{kang2019decoupling}
Bingyi Kang, Saining Xie, Marcus Rohrbach, Zhicheng Yan, Albert Gordo, Jiashi Feng, and Yannis Kalantidis.
\newblock Decoupling representation and classifier for long-tailed recognition.
\newblock \emph{arXiv preprint arXiv:1910.09217}, 2019.

\bibitem[Kinoshita et~al.(2024)Kinoshita, Maruyama, Kobayashi, Imanishi, Maruyama, Ohira, Endo, Tochigi, Kinoshita, Fukui, et~al.]{kinoshita2024artificial}
Kazuya Kinoshita, Tetsuro Maruyama, Nao Kobayashi, Shunsuke Imanishi, Michihiro Maruyama, Gaku Ohira, Satoshi Endo, Toru Tochigi, Mayuko Kinoshita, Yudai Fukui, et~al.
\newblock An artificial intelligence-based nerve recognition model is useful as surgical support technology and as an educational tool in laparoscopic and robot-assisted rectal cancer surgery.
\newblock \emph{Surgical Endoscopy}, 38\penalty0 (9):\penalty0 5394--5404, 2024.

\bibitem[Liu et~al.(2023)Liu, Li, Yao, Fan, Hu, Dawant, Nath, Xu, and Oguz]{liu2023colossal}
Han Liu, Hao Li, Xing Yao, Yubo Fan, Dewei Hu, Benoit~M Dawant, Vishwesh Nath, Zhoubing Xu, and Ipek Oguz.
\newblock {COLosSAL}: A benchmark for cold-start active learning for {3D} medical image segmentation.
\newblock In \emph{International Conference on Medical Image Computing and Computer-Assisted Intervention}, pages 25--34. Springer, 2023.

\bibitem[Liu et~al.(2025)Liu, Yin, Jing, Zhu, Chen, Xi, Feng, Wang, Li, and Shen]{liu2025cosine}
Yang Liu, Yufei Yin, Chenchen Jing, Muzhi Zhu, Hao Chen, Yuling Xi, Bo Feng, Hao Wang, Shiyu Li, and Chunhua Shen.
\newblock Unified open-world segmentation with multi-modal prompts.
\newblock In \emph{2025 IEEE/CVF International Conference on Computer Vision (ICCV)}, pages 21557--21567. IEEE, 2025.

\bibitem[Ma et~al.(2024)Ma, He, Li, Han, You, and Wang]{ma2024medsam}
Jun Ma, Yuting He, Feifei Li, Lin Han, Chenyu You, and Bo Wang.
\newblock Segment anything in medical images.
\newblock \emph{Nature communications}, 15\penalty0 (1):\penalty0 654, 2024.

\bibitem[Ma et~al.(2025)Ma, Yang, Kim, Chen, Baharoon, Fallahpour, Asakereh, Lyu, and Wang]{ma2025medsam2}
Jun Ma, Zongxin Yang, Sumin Kim, Bihui Chen, Mohammed Baharoon, Adibvafa Fallahpour, Reza Asakereh, Hongwei Lyu, and Bo Wang.
\newblock {MedSAM2}: Segment anything in {3D} medical images and videos.
\newblock \emph{arXiv preprint arXiv:2504.03600}, 2025.

\bibitem[Madani et~al.(2022)Madani, Namazi, Altieri, Hashimoto, Rivera, Pucher, Navarrete-Welton, Sankaranarayanan, Brunt, Okrainec, et~al.]{madani2022artificial}
Amin Madani, Babak Namazi, Maria~S Altieri, Daniel~A Hashimoto, Angela~Maria Rivera, Philip~H Pucher, Allison Navarrete-Welton, Ganesh Sankaranarayanan, L~Michael Brunt, Allan Okrainec, et~al.
\newblock Artificial intelligence for intraoperative guidance: using semantic segmentation to identify surgical anatomy during laparoscopic cholecystectomy.
\newblock \emph{Annals of surgery}, 276\penalty0 (2):\penalty0 363--369, 2022.

\bibitem[Maier-Hein et~al.(2022)Maier-Hein, Eisenmann, Sarikaya, M{\"a}rz, Collins, Malpani, Fallert, Feussner, Giannarou, Mascagni, et~al.]{maierhein2022surgical}
Lena Maier-Hein, Matthias Eisenmann, Duygu Sarikaya, Keno M{\"a}rz, Toby Collins, Anand Malpani, Johannes Fallert, Hubertus Feussner, Stamatia Giannarou, Pietro Mascagni, et~al.
\newblock Surgical data science--from concepts toward clinical translation.
\newblock \emph{Medical image analysis}, 76:\penalty0 102306, 2022.

\bibitem[Mascagni et~al.(2025)Mascagni, Alapatt, Murali, et~al.]{mascagni2025endoscapes}
Pietro Mascagni, Deepak Alapatt, Aditya Murali, et~al.
\newblock {Endoscapes}, a critical view of safety and surgical scene segmentation dataset for laparoscopic cholecystectomy.
\newblock \emph{Scientific Data}, 12:\penalty0 331, 2025.

\bibitem[Moreu et~al.(2022)Moreu, McGuinness, and O'Connor]{moreu2022synthetic}
Enric Moreu, Kevin McGuinness, and Noel~E. O'Connor.
\newblock Synthetic data for unsupervised polyp segmentation.
\newblock \emph{arXiv preprint arXiv:2202.08680}, 2022.

\bibitem[Mousavi et~al.(2025)Mousavi, Ozbulak, Tozzi, Rashidian, Willaert, Vankerschaver, and De~Neve]{mousavi2025one}
Seyed~Amir Mousavi, Utku Ozbulak, Francesca Tozzi, Nikdokht Rashidian, Wouter Willaert, Joris Vankerschaver, and Wesley De~Neve.
\newblock One patient's annotation is another one's initialization: Towards zero-shot surgical video segmentation with cross-patient initialization.
\newblock \emph{arXiv preprint arXiv:2503.02228}, 2025.

\bibitem[Nemhauser et~al.(1978)Nemhauser, Wolsey, and Fisher]{nemhauser1978analysis}
George~L Nemhauser, Laurence~A Wolsey, and Marshall~L Fisher.
\newblock An analysis of approximations for maximizing submodular set functions---{I}.
\newblock \emph{Mathematical programming}, 14\penalty0 (1):\penalty0 265--294, 1978.

\bibitem[Ninomiya et~al.(2026)Ninomiya, Matsunobu, Oyama, Takayama, Masuda, Hirashita, Endo, Tokuyasu, and Inomata]{ninomiya2026educational}
Shigeo Ninomiya, Yusuke Matsunobu, Tabito Oyama, Hiroomi Takayama, Takashi Masuda, Teijiro Hirashita, Yuichi Endo, Tatsushi Tokuyasu, and Masafumi Inomata.
\newblock Educational impact of artificial intelligence-navigation surgery on anatomical landmark recognition in medical students.
\newblock \emph{Annals of Gastroenterological Surgery}, 10\penalty0 (3):\penalty0 843--850, 2026.

\bibitem[Oh et~al.(2019)Oh, Lee, Xu, and Kim]{oh2019video}
Seoung~Wug Oh, Joon-Young Lee, Ning Xu, and Seon~Joo Kim.
\newblock Video object segmentation using space-time memory networks.
\newblock In \emph{2019 IEEE/CVF International Conference on Computer Vision (ICCV)}, pages 9225--9234. IEEE, 2019.

\bibitem[Protserov et~al.(2024)Protserov, Hunter, Zhang, Mashouri, Masino, Brudno, and Madani]{protserov2024development}
Sergey Protserov, Jaryd Hunter, Haochi Zhang, Pouria Mashouri, Caterina Masino, Michael Brudno, and Amin Madani.
\newblock Development, deployment and scaling of operating room-ready artificial intelligence for real-time surgical decision support.
\newblock \emph{NPJ Digital Medicine}, 7\penalty0 (1):\penalty0 231, 2024.

\bibitem[Psychogyios et~al.(2023)Psychogyios, Colleoni, Van~Amsterdam, Li, Huang, Li, Jia, Zou, Wang, Liu, et~al.]{psychogyios2023sarrarp50}
Dimitrios Psychogyios, Emanuele Colleoni, Beatrice Van~Amsterdam, Chih-Yang Li, Shu-Yu Huang, Yuchong Li, Fucang Jia, Baosheng Zou, Guotai Wang, Yang Liu, et~al.
\newblock {SAR-RARP50}: Segmentation of surgical instrumentation and action recognition on robot-assisted radical prostatectomy challenge.
\newblock \emph{arXiv preprint arXiv:2401.00496}, 2023.

\bibitem[Qi et~al.(2025)Qi, Zhu, Li, Bi, Qi, Ma, and Yang]{qi2025dcsam}
Mengshi Qi, Pengfei Zhu, Xiangtai Li, Xiaoyang Bi, Lu Qi, Huadong Ma, and Ming-Hsuan Yang.
\newblock {DC-SAM}: In-context segment anything in images and videos via dual consistency.
\newblock \emph{IEEE Transactions on Pattern Analysis and Machine Intelligence}, 2025.

\bibitem[Rakic et~al.(2024)Rakic, Wong, Ortiz, Cimini, Guttag, and Dalca]{rakic2024tyche}
Marianne Rakic, Hallee~E Wong, Jose Javier~Gonzalez Ortiz, Beth~A Cimini, John~V Guttag, and Adrian~V Dalca.
\newblock {Tyche}: Stochastic in-context learning for medical image segmentation.
\newblock In \emph{2024 IEEE/CVF Conference on Computer Vision and Pattern Recognition (CVPR)}, pages 11159--11173. IEEE, 2024.

\bibitem[Ravi et~al.(2025)Ravi, Gabeur, Hu, Hu, Ryali, Ma, Khedr, R{\"a}dle, Rolland, Gustafson, et~al.]{ravi2025sam2}
Nikhila Ravi, Valentin Gabeur, Yuan-Ting Hu, Ronghang Hu, Chaitanya Ryali, Tengyu Ma, Haitham Khedr, Roman R{\"a}dle, Chloe Rolland, Laura Gustafson, et~al.
\newblock {SAM}~2: Segment anything in images and videos.
\newblock In \emph{International Conference on Learning Representations}, pages 28085--28128, 2025.

\bibitem[Ronneberger et~al.(2015)Ronneberger, Fischer, and Brox]{ronneberger2015unet}
Olaf Ronneberger, Philipp Fischer, and Thomas Brox.
\newblock {U-Net}: Convolutional networks for biomedical image segmentation.
\newblock In \emph{International Conference on Medical image computing and computer-assisted intervention}, pages 234--241. Springer, 2015.

\bibitem[Sapkota et~al.(2026)Sapkota, Zhang, Zheng, Ma, Shi, Mariam, and Chen]{sapkota2026dgem}
Nishchal Sapkota, Yejia Zhang, Bofang Zheng, Xianshi Ma, Haoyan Shi, Yohannes Mariam, and Danny~Z Chen.
\newblock Sparsely supervised surgical video segmentation with reliable asymmetric dual memory.
\newblock In \emph{International Conference on Medical Image Computing and Computer-Assisted Intervention}, pages 509--519. Springer, 2026.

\bibitem[Scheikl et~al.(2020)Scheikl, Laschewski, Kisilenko, Davitashvili, M{\"u}ller, Capek, M{\"u}ller-Stich, Wagner, and Mathis-Ullrich]{scheikl2020deep}
Paul~Maria Scheikl, Stefan Laschewski, Anna Kisilenko, Tornike Davitashvili, Benjamin M{\"u}ller, Manuela Capek, Beat~P M{\"u}ller-Stich, Martin Wagner, and Franziska Mathis-Ullrich.
\newblock Deep learning for semantic segmentation of organs and tissues in laparoscopic surgery.
\newblock \emph{Current Directions in Biomedical Engineering}, 6\penalty0 (1):\penalty0 20200016, 2020.

\bibitem[Sener and Savarese(2017)]{sener2018coreset}
Ozan Sener and Silvio Savarese.
\newblock Active learning for convolutional neural networks: A core-set approach.
\newblock \emph{arXiv preprint arXiv:1708.00489}, 2017.

\bibitem[Sim{\'e}oni et~al.(2025)Sim{\'e}oni, Vo, Seitzer, Baldassarre, Oquab, Jose, Khalidov, Szafraniec, Yi, Ramamonjisoa, et~al.]{simeoni2026dinov3}
Oriane Sim{\'e}oni, Huy~V Vo, Maximilian Seitzer, Federico Baldassarre, Maxime Oquab, Cijo Jose, Vasil Khalidov, Marc Szafraniec, Seungeun Yi, Micha{\"e}l Ramamonjisoa, et~al.
\newblock {DINOv3}.
\newblock \emph{arXiv preprint arXiv:2508.10104}, 2025.

\bibitem[Skinner et~al.(2024)Skinner, Chen, Jentis, Liu, McCulloh, Harzman, Huang, Kalady, and Kim]{skinner2024realtime}
Garrett Skinner, Tina Chen, Gabriel Jentis, Yao Liu, Christopher McCulloh, Alan Harzman, Emily Huang, Matthew Kalady, and Peter Kim.
\newblock Real-time near infrared artificial intelligence using scalable non-expert crowdsourcing in colorectal surgery.
\newblock \emph{NPJ Digital Medicine}, 7\penalty0 (1):\penalty0 99, 2024.

\bibitem[Su et~al.(2022)Su, Kasai, Wu, Shi, Wang, Xin, Zhang, Ostendorf, Zettlemoyer, Smith, et~al.]{su2023selective}
Hongjin Su, Jungo Kasai, Chen~Henry Wu, Weijia Shi, Tianlu Wang, Jiayi Xin, Rui Zhang, Mari Ostendorf, Luke Zettlemoyer, Noah~A Smith, et~al.
\newblock Selective annotation makes language models better few-shot learners.
\newblock \emph{arXiv preprint arXiv:2209.01975}, 2022.

\bibitem[Sun et~al.(2026)Sun, Ren, and Zhang]{sun2026rsrm}
Sujun Sun, Mingwu Ren, and Haofeng Zhang.
\newblock Training-free cross-domain few-shot segmentation via robust semantic representation and matching.
\newblock In \emph{European Conference on Computer Vision}, pages 370--388. Springer, 2026.

\bibitem[Suo et~al.(2024)Suo, Lai, Sun, Zhang, Wang, and Zhang]{suo2024rethinking}
Wei Suo, Lanqing Lai, Mengyang Sun, Hanwang Zhang, Peng Wang, and Yanning Zhang.
\newblock Rethinking and improving visual prompt selection for in-context learning segmentation.
\newblock In \emph{European Conference on Computer Vision}, pages 18--35. Springer, 2024.

\bibitem[Twinanda et~al.(2017)Twinanda, Shehata, Mutter, Marescaux, De~Mathelin, and Padoy]{twinanda2016endonet}
Andru~P Twinanda, Sherif Shehata, Didier Mutter, Jacques Marescaux, Michel De~Mathelin, and Nicolas Padoy.
\newblock {EndoNet}: a deep architecture for recognition tasks on laparoscopic videos.
\newblock \emph{IEEE transactions on medical imaging}, 36\penalty0 (1):\penalty0 86--97, 2017.

\bibitem[Videnovic et~al.(2025)Videnovic, Lukezic, and Kristan]{videnovic2025distractor}
Jovana Videnovic, Alan Lukezic, and Matej Kristan.
\newblock A distractor-aware memory for visual object tracking with {SAM2}.
\newblock In \emph{2025 IEEE/CVF Conference on Computer Vision and Pattern Recognition (CVPR)}, pages 24255--24264. IEEE, 2025.

\bibitem[Wahd et~al.(2025)Wahd, Jaremko, and Hareendranathan]{wahd2025temporal}
Assefa Wahd, Jacob Jaremko, and Abhilash Hareendranathan.
\newblock Time-contrastive pretraining for in-context image and video segmentation.
\newblock In \emph{International Conference on Medical Image Computing and Computer-Assisted Intervention}, pages 630--639. Springer, 2025.

\bibitem[Wang et~al.(2023{\natexlab{a}})Wang, Wang, Cao, Shen, and Huang]{wang2023painter}
Xinlong Wang, Wen Wang, Yue Cao, Chunhua Shen, and Tiejun Huang.
\newblock Images speak in images: A generalist painter for in-context visual learning.
\newblock In \emph{2023 IEEE/CVF Conference on Computer Vision and Pattern Recognition (CVPR)}, pages 6830--6839. IEEE, 2023{\natexlab{a}}.

\bibitem[Wang et~al.(2023{\natexlab{b}})Wang, Zhang, Cao, Wang, Shen, and Huang]{wang2023seggpt}
Xinlong Wang, Xiaosong Zhang, Yue Cao, Wen Wang, Chunhua Shen, and Tiejun Huang.
\newblock {SegGPT}: Towards segmenting everything in context.
\newblock In \emph{2023 IEEE/CVF International Conference on Computer Vision (ICCV)}, pages 1130--1140. IEEE, 2023{\natexlab{b}}.

\bibitem[Ward et~al.(2021)Ward, Mascagni, Madani, Padoy, Perretta, and Hashimoto]{ward2021surgical}
Thomas~M Ward, Pietro Mascagni, Amin Madani, Nicolas Padoy, Silvana Perretta, and Daniel~A Hashimoto.
\newblock Surgical data science and artificial intelligence for surgical education.
\newblock \emph{Journal of Surgical Oncology}, 124\penalty0 (2):\penalty0 221--230, 2021.

\bibitem[Wei et~al.(2015)Wei, Iyer, and Bilmes]{wei2015submodularity}
Kai Wei, Rishabh Iyer, and Jeff Bilmes.
\newblock Submodularity in data subset selection and active learning.
\newblock In \emph{International conference on machine learning}, pages 1954--1963. PMLR, 2015.

\bibitem[Wu et~al.(2024)Wu, Restrepo, Shuai, Liu, and Shen]{wu2024efficient}
Chenwei Wu, David Restrepo, Zitao Shuai, Zhongming Liu, and Liyue Shen.
\newblock Efficient in-context medical segmentation with meta-driven visual prompt selection.
\newblock In \emph{International Conference on Medical Image Computing and Computer-Assisted Intervention}, pages 255--265. Springer, 2024.

\bibitem[Yang et~al.(2024)Yang, Huang, Chai, Jiang, and Hwang]{yang2024samurai}
Cheng-Yen Yang, Hsiang-Wei Huang, Wenhao Chai, Zhongyu Jiang, and Jenq-Neng Hwang.
\newblock {SAMURAI}: Adapting segment anything model for zero-shot visual tracking with motion-aware memory.
\newblock \emph{arXiv preprint arXiv:2411.11922}, 2024.

\bibitem[Yin et~al.(2021)Yin, Zheng, Luo, Qian, Zhang, and Gao]{yin2021learning}
Zhaoyuan Yin, Jia Zheng, Weixin Luo, Shenhan Qian, Hanling Zhang, and Shenghua Gao.
\newblock Learning to recommend frame for interactive video object segmentation in the wild.
\newblock In \emph{2021 IEEE/CVF Conference on Computer Vision and Pattern Recognition (CVPR)}, pages 15440--15449. IEEE, 2021.

\bibitem[Yoon et~al.(2022)Yoon, Hong, Hong, Lee, Shin, Park, Sung, Yu, Kim, Park, et~al.]{yoon2022surgical}
Jihun Yoon, SeulGi Hong, Seungbum Hong, Jiwon Lee, Soyeon Shin, Bokyung Park, Nakjun Sung, Hayeong Yu, Sungjae Kim, SungHyun Park, et~al.
\newblock Surgical scene segmentation using semantic image synthesis with a virtual surgery environment.
\newblock In \emph{International Conference on Medical Image Computing and Computer-Assisted Intervention}, pages 551--561. Springer, 2022.

\bibitem[Yue et~al.(2024)Yue, Zhang, Hu, Xia, Luo, and Wang]{yue2024surgicalsam}
Wenxi Yue, Jing Zhang, Kun Hu, Yong Xia, Jiebo Luo, and Zhiyong Wang.
\newblock {SurgicalSAM}: Efficient class promptable surgical instrument segmentation.
\newblock In \emph{Proceedings of the AAAI Conference on Artificial Intelligence}, pages 6890--6898, 2024.

\bibitem[Zhang et~al.(2024)Zhang, Jiang, Guo, Yan, Pan, Dong, Qiao, Peng, and Li]{zhang2024persam}
Renrui Zhang, Zhengkai Jiang, Ziyu Guo, Shilin Yan, Junting Pan, Hao Dong, Yu Qiao, Gao Peng, and Hongsheng Li.
\newblock Personalize segment anything model with one shot.
\newblock In \emph{International Conference on Learning Representations}, pages 18250--18279, 2024.

\bibitem[Zhang et~al.(2023)Zhang, Zhou, and Liu]{zhang2023makes}
Yuanhan Zhang, Kaiyang Zhou, and Ziwei Liu.
\newblock What makes good examples for visual in-context learning?
\newblock \emph{Advances in Neural Information Processing Systems}, 36:\penalty0 17773--17794, 2023.

\bibitem[Zheng et~al.(2019)Zheng, Yang, Chen, Han, Zhang, Liang, Zhao, Wang, and Chen]{zheng2019representative}
Hao Zheng, Lin Yang, Jianxu Chen, Jun Han, Yizhe Zhang, Peixian Liang, Zhuo Zhao, Chaoli Wang, and Danny~Z Chen.
\newblock Biomedical image segmentation via representative annotation.
\newblock In \emph{Proceedings of the AAAI Conference on Artificial Intelligence}, pages 5901--5908, 2019.

\bibitem[Zhou et~al.(2003)Zhou, Bousquet, Lal, Weston, and Sch{\"o}lkopf]{zhou2003learning}
Dengyong Zhou, Olivier Bousquet, Thomas Lal, Jason Weston, and Bernhard Sch{\"o}lkopf.
\newblock Learning with local and global consistency.
\newblock \emph{Advances in neural information processing systems}, 16, 2003.

\end{thebibliography}
